\documentclass[conference]{IEEEtran}
\IEEEoverridecommandlockouts
\usepackage{cite}
\usepackage{color}
\usepackage{amsmath,amssymb,amsfonts}
\usepackage{algorithmic}
\usepackage{graphicx}
\usepackage{textcomp}
\usepackage{bm} 
\usepackage{multirow}
\usepackage{booktabs}
\usepackage{hyperref}
\usepackage{cleveref}
\usepackage{fancyvrb}
\usepackage{subcaption}
\usepackage{listings}
\usepackage{xcolor} 
\usepackage{colortbl}
\usepackage[most]{tcolorbox} 
\tcbset{breakable}
\usepackage{listings} 
\usepackage[numbers, sort&compress]{natbib}
\hypersetup{
    colorlinks=true,
    linkcolor=black,
    citecolor=black,
    urlcolor=black,
}
\newtcolorbox{promptbox}[1]{
  enhanced,
  colback=green!5!gray!10,
  colframe=gray!50!black,
  arc=0.5mm,
  boxrule=1pt,
  title=#1,
  fonttitle=\bfseries\color{white},
  fontlower=\rmfamily
  coltitle=gray!50!black,
  breakable=false,
  width=\linewidth
}

\def\BibTeX{{\rm B\kern-.05em{\sc i\kern-.025em b}\kern-.08em
    T\kern-.1667em\lower.7ex\hbox{E}\kern-.125emX}}

\newcommand{\std}[1]{\fontsize{5}{7}\selectfont $\pm${#1}}

\begin{document}

\title{Augmenting Continual Learning of Diseases with LLM-Generated Visual Concepts}

\author {
    Jiantao~Tan\textsuperscript{\rm 1}, Peixian~Ma\textsuperscript{\rm 2}, Kanghao Chen\textsuperscript{\rm 2}, Zhiming~Dai\textsuperscript{\rm 1}, Ruixuan Wang\textsuperscript{\rm 1,3,4}\\
    \\
    \textsuperscript{\rm 1}Sun Yat-sen University, Guangzhou, China \\
    \textsuperscript{\rm 2}The Hong Kong University of Science and Technology (Guangzhou), Guangzhou, China \\
    \textsuperscript{\rm 3}Peng Cheng Laboratory, Shenzhen, China \\
    \textsuperscript{\rm 4}Key Laboratory of Machine Intelligence and Advanced Computing, MOE, Guangzhou, China
}

\maketitle

\begin{abstract}
Continual learning is essential for medical image classification systems to adapt to dynamically evolving clinical environments. The integration of multimodal information can significantly enhance continual learning of image classes. However, while existing approaches do utilize textual modality information, they solely rely on simplistic templates with a class name, thereby neglecting richer semantic information. To address these limitations, we propose a novel framework that harnesses visual concepts generated by large language models (LLMs) as discriminative semantic guidance. Our method dynamically constructs a visual concept pool with a similarity-based filtering mechanism to prevent redundancy. Then, to integrate the concepts into the continual learning process, we employ a cross-modal image-concept attention module, coupled with an attention loss. Through attention, the module can leverage the semantic knowledge from relevant visual concepts and produce  class-representative fused features for classification. Experiments on medical and natural image datasets show our method achieves state-of-the-art performance, demonstrating the effectiveness and superiority of our method. We will release the code publicly.
\end{abstract}


\section{Introduction}

With the widespread applications of deep learning techniques in medical imaging, computer-aided diagnosis systems have demonstrated significant potential in disease diagnosis. 
However, while most existing deep learning models excel in specific tasks, they struggle to address the challenges posed by the continuous emergence of new disease types in clinical scenarios. When updating the model with data of new diseases, traditional retraining approaches face catastrophic forgetting~\cite{catastrophic-forgetting}. Besides, due to the privacy constraints inherent in medical data, the model is often prevented from storing or retrieving old data, which further leads to performance degradation in old classes.

\begin{figure}
    \centering
    \includegraphics[width=\linewidth]{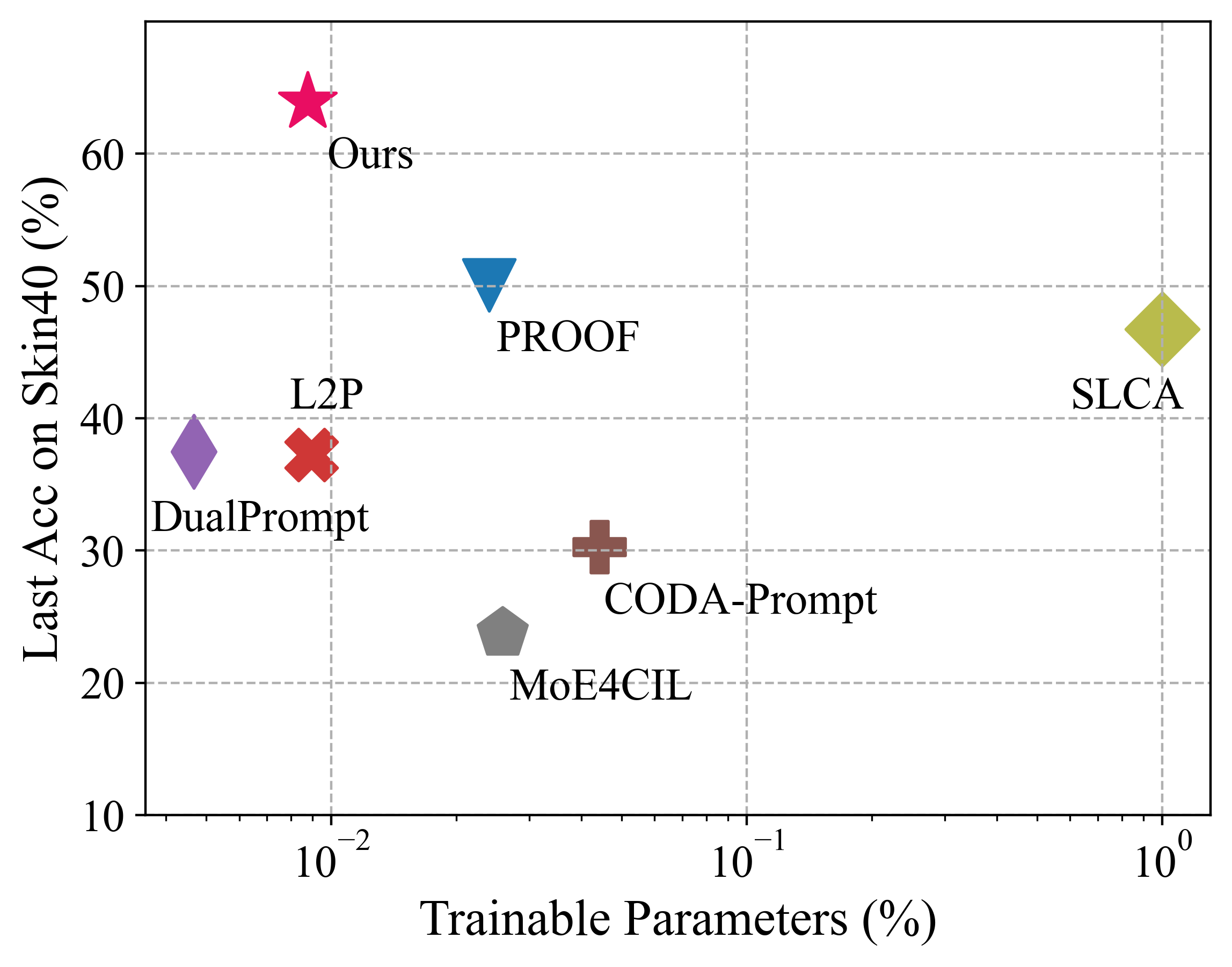}
    \caption{Comparison of our method with strong baselines on the Skin40 dataset in the 10-task setting. }
    \label{fig:param_skin}
\end{figure}

In recent years, the challenge of catastrophic forgetting has prompted significant advances in continual learning (CL). Currently, research efforts have extended CL techniques~\cite{TexCIL, ACL_structure, zhao2023class} to medical imaging domains, achieving great success by adapting methodologies from the natural image domain. 
With the emergence of multimodal models, particularly vision-language models (\textit{e.g.}, CLIP~\cite{CLIP}), recent studies have incorporated textual information into image CL frameworks~\cite{PROOF, MoE-adapters, CBM}, yielding superior performance compared to previous unimodal approaches. This advancement stems from the rich semantic information inherently carried by language, which provides high-level guidance for visual learning and enables the acquisition of discriminative and interpretable class features. This approach has also been demonstrated to be effective in the medical image domain. However, most existing studies overlook detailed semantic information on textual modality, instead relying only on a simple template-based text with the class name that lacks comprehensive semantic knowledge~\cite {TexCIL, PROOF}. 


\begin{figure*}[!t]
    \centering
    \includegraphics[width=0.99\linewidth]{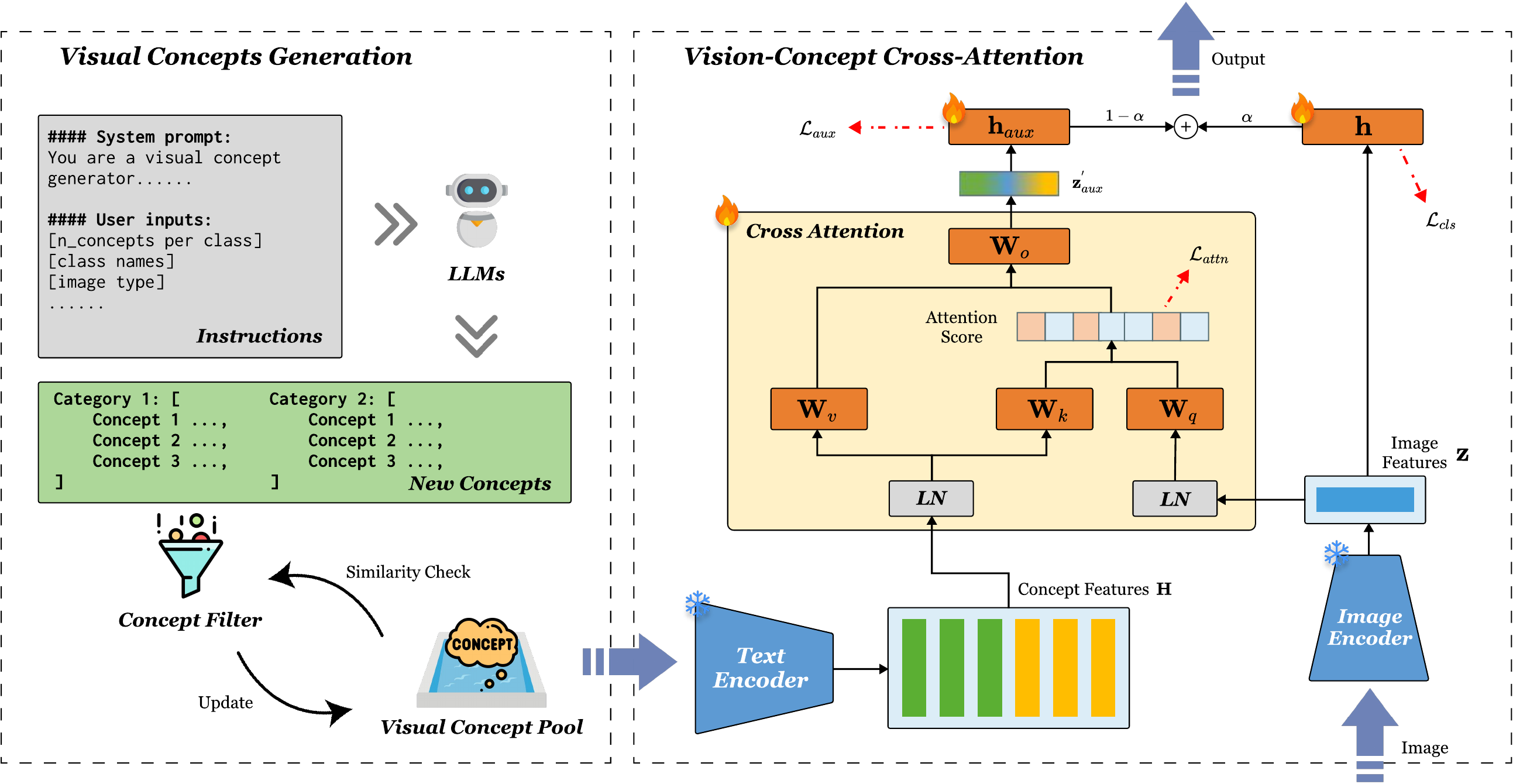}
    \caption{Illustration of the proposed framework. Left: the proposed concept generation pipeline. For novel classes, new concepts are generated by LLM, selected through a filtering mechanism, and then updated into the concept pool. Right: the proposed vision-concept cross-attention module. Image features interact with concept features via cross-attention to obtain fused features, which are subsequently passed through an auxiliary classifier head. The final output is obtained through a weighted sum.}
    \label{fig:method}
\end{figure*}

Intuitively, human inherently associates detailed visual semantic information with linguistic representations during visual recognition, deconstructing complex images as combinations of visual concepts (\textit{e.g.}, the term \textit{``zebra''} evokes \textit{``black-and-white stripes''}). This mechanism can be integrated into continual learning frameworks.
Critically, these concepts function as static anchors with concentrated semantic information, providing robust class discriminative cues that mitigate catastrophic forgetting during continual model updates. Furthermore, the emergence of large language models (LLMs) with extensive knowledge repositories and strong generalization capabilities has offered widespread applications across domains~\cite{sqlr1}. Therefore, LLMs can be used to provide diverse semantic textual priors to support CL tasks.

Building upon these insights, this study introduces a novel exemplar-free CL framework that enhances continual classification performance by incorporating vision and concept text fusion. Specifically, we maintain a visual concept pool 
generated by LLM. The concept pool is dynamically updated via a similarity-based concept filtering mechanism, which can prevent concept overlap between confusable classes. The visual concepts serve as static discriminative priors that remain stable and help enhance the robustness of the model to forgetting. To utilize these concepts in CL process, we design a cross-modal image-concept cross-attention module for image-text interaction. This module uses the image feature to query and fuse relevant visual concepts, obtaining the class-discriminative fused feature. An auxiliary classifier processes this fused feature, and the output can enhance the image classification result through a weighted combination. Furthermore, an attention constraint loss is introduced to enforce image features to focus on corresponding concepts while suppressing interference from irrelevant concepts, thereby producing a more semantic-focused fused feature. In summary, our contributions are as follows.
\begin{itemize}
    \item We introduce visual concepts to assist in the continual learning of images, which can produce a discriminative feature and enhance the model's resistance to forgetting;
    \item We propose a cross-modal interaction mechanism between the image feature and visual concepts through a cross-attention module;
    \item Our method achieves state-of-the-art performance on multiple CL benchmarks.
\end{itemize}

\section{Related Work}

\subsection{Class-Incremental Learning}
Class Incremental Learning (CIL) aims to enable learning systems to gradually acquire knowledge from data of new classes over time. Approaches to CIL can generally be categorized into four types. Parameter regularization methods~\cite{EWC} alleviate the problem of catastrophic forgetting by imposing regularization constraints on the model parameters. These methods prevent the model from forgetting previously learned knowledge when learning new classes by constraining changes in model parameters. Knowledge distillation-based methods~\cite{LwF, WA, UCIR} utilize knowledge distillation techniques to regulate the model's outputs, helping the current model mimic the behavior of the previous model during the updating process. This approach reduces forgetting by transferring the knowledge from the old model to the new model. Replay-based methods~\cite{iCarL, BiC, DarkER} typically maintain a memory buffer that stores a small portion of old data for experience replay. This approach helps the model retain previously learned knowledge by re-accessing old data.
Replay-based methods are often combined with knowledge distillation to perform experience replay in feature or label space, demonstrating excellent performance. 
Structure-based methods~\cite{ACL_structure, DER, TexCIL} typically introduce new learnable modules to facilitate knowledge acquisition from new data, thus avoiding parameter interference and task conflicts.

\subsection{Class-Incremental Learning on Pretrained Model}
Due to the strong generalization capabilities of pre-trained models, recent research~\cite{SLCA, CBM, zhao2023class} has focused on implementing continuous learning (CL) using such models. Applying pre-trained models to downstream continuous learning tasks typically involves parameter-efficient fine-tuning (PEFT) techniques~\cite{peft}. These methods include prompt-based continual fine-tuning approaches~\cite{L2P, DualPrompt, CODA-Prompt} as well as adapter fine-tuning methods~\cite{EASE, SSIAT}. 
Furthermore, some studies have explored continual learning through pre-trained vision-language models (\textit{e.g.}, CLIP~\cite{CLIP}) by fine-tuning the image encoder~\cite{MoE-adapters, TexCIL}. Others try to mitigate forgetting through cross-modal interaction based on pre-trained image features~\cite{PROOF, CBM}. However, while these approaches leverage information from the text side of VLMs, they tend to remain focused on simple class texts, lacking the extraction of richer semantic information for assisted learning. In contrast, this paper leverages large models to generate more semantically rich and nuanced conceptual texts for visual concepts, thereby guiding models to maintain a more stable class discrimination capability throughout the continual learning process.

\section{Methodology}
\subsection{Problem Formulation}
In this study, we address the challenge of class-incremental learning. The class-incremental learning process is organized into $T$ tasks, with each task $t \in \{1, 2, \ldots, T\}$ corresponding to a distinct subset of data classes $\mathcal{C}_t \subset \mathcal{C}$, where $\mathcal{C}$ denotes the complete set of data classes. These subsets satisfy the conditions $\mathcal{C}_i \cap \mathcal{C}_j = \emptyset$ for all $i \neq j$, and the union of all class sets across the tasks composes the complete set (\textit{i.e.}, $\bigcup \mathcal{C}_t = \mathcal{C}$). The primary objective of class-incremental learning is to develop a classifier that can precisely determine the category of an input image belonging to $\mathcal{C}$ after finishing all tasks' learning. Additionally, this study focuses on the realistic scenario in which real training data from old classes are unavailable. 

\begin{figure}[!t]
    \centering
    \begin{promptbox}{Instruction and output example}
\textbf{Instruction:}\\
You are a visual concept generator. Your task is to generate concise visual concepts that are the most representative of the input class (\textit{e.g.}, ``black
and white stripes'' for a ``zebra''). \\
\\
\textit{Inputs:} \\
- Number of concepts per class: [n\_concepts] \\
- Class name: [class\_name] \\
- Image types: [Natural/Dermoscopy/...]\\
\\
\textit{Output Format:} \\
Return a JSON object: \\
\{`class\_name1':[`concept1',`concept2',`concept3'], `class\_name2':[...], ...\} \\
\\
\textit{Rules:}\\
1. Focus on distinctive visual features of the class;\\
2. Use descriptive adjectives + nouns format;\\
3. Avoid non-visual features;\\
4. Each concept should be distinct from others within the same class.\\
\\
\textbf{Example of output:}\\
  \textit{`Onychomycosis'}: [\\
  `thickened discolored nail', \\
  `brittle crumbling edges', \\
  `yellowish nail streaks' 
  ]

    \end{promptbox}
    \caption{The prompt template and output example of LLM.}
    \label{fig:prompt}
\end{figure}

\subsection{Framework Overview}
To address the CIL challenge, a novel concept-guided class-incremental learning framework is proposed, as illustrated in \Cref{fig:method}. 
The framework's core innovation involves leveraging visual concepts generated by a Large Language Model (LLM) as semantic signals to improve class discriminability and mitigate forgetting. It comprises two primary components. The first component includes a dynamic visual concept pool, wherein the LLM generates diverse visual concept descriptions for new classes (\textit{e.g.}, concept \textit{``long neck"} and concept \textit{``spots"} for class \textit{``giraffe"}), paired with a semantic similarity-driven concept filtering mechanism that can prevent concept redundancy among easily confusable classes. The second component is the image-concept interaction mechanism, comprising a vision-concept cross-attention module.
This cross-attention mechanism can produce a semantically concentrated fused feature, which serves as more robust static discriminative information to assist classification during the CL process. To further refine attention, an attention constraint loss is imposed to emphasize key concepts relevant to the image's class while suppressing attention to irrelevant concepts. Ultimately, the resulting fused feature is processed through an auxiliary classifier head to improve the classification performance by output weighting.

\subsection{Visual Concepts Generation}
Inspired by human cognition, where learning new image classes inherently involves constructing unique visual features through language, our proposed method employs an LLM 
to generate visual concepts for each class, which are then integrated into the continual learning process as semantic linguistic information. To this end, we introduce a dynamically maintained visual concept pool ($P_{vc}$). At each incremental learning task, the large language model generates class-specific visual concepts which are selectively written into $P_{vc}$. 
Specifically, for the current learning task $ t $, the LLM is utilized to produce a set $P_t$ of visual concept texts for each class $ c \in \mathcal{C}_t $:
\begin{align}
    P_t=LLM(\mathcal{C}_t, k, \Gamma) \,,
\end{align}
where $\Gamma$ is the input template as shown in \Cref{fig:prompt} and $k$ is the number of concepts generated for each class. There may be confusion among the classes of the new and old data, especially for fine-grained classification tasks, which can be manifested by the overlap or semantic similarity of visual concepts. Therefore, we design a filtering mechanism based on TF-IDF (Term Frequency-Inverse Document Frequency) to update $P_{vc}$. For each new concept $d_{new} $, we calculate the similarities between the TF-IDF vectors of $d_{new}$ and all concepts $\{d_i\}$ in the visual concept pool $P_ {vc}$, denoted as:
\begin{equation}
   \begin{aligned}
        S = \{s_i | s_i = cos(\mathbf{v}_{new}, \mathbf{v}_i), \forall i=1,...,|P_{vc}|\}~,
    \end{aligned} 
\end{equation}
where $cos$ denotes the cosine similarity measurement of two vectors, and $\mathbf{v}_{new}$ and $\mathbf{v}_{i}$ are the TF-IDF vectors for concepts $d_{new}$ and $d_i$, respectively.  $\mathbf{v}_{new}$ is computed based on a set $\mathcal{V}$ which is the collection of all $n$-gram terms appearing in the two concepts $d_{new}$ and $d_i$, with $n$ varying from 1 to the maximum number of words in the two concepts.
For example, \textit{``black and white''} is a 3-gram term of the concept \textit{``black and white stripes''} in $\mathcal{V}$. Suppose there are totally $M$ terms in $\mathcal{V}$. 
The vector $\mathbf{v}_{new} = [tfidf(w_1), tfidf(w_2), \ldots, tfidf(w_M)]$ is obtained by computing the TF-IDF values for all the $M$ terms $\{w_m, m=1,\ldots, M\}$ \cite{tfidf}. Similarly, $\mathbf{v}_i$ can be computed for concept $d_i$.
If $max(S)$ is less that threshold $\tau$, then $d_{new}$ is considered dissimilar to all concepts in $P_{vc}$ and is added as a new concept of the new class; otherwise, $d_{new}$ is replaced with the existing concept corresponding to $max(S)$.

\subsection{Vision-Concept Cross-Attention}

Based on the acquisition of the visual concept pool, we devise a cross-modal cross-attention module designed to retrieve and integrate visual concepts associated with the input image, thereby generating semantic-condensed fused feature. Specifically, the image features $\mathbf{z}$, which are derived from the input image $\mathbf{x}$ by passing it through the image encoder $f(\cdot)$, acts as the query and the textual features of the visual concepts serve as the keys and values in the cross-attention process. This approach allows the model to leverage the concept features that are relevant to the image, thereby producing a discriminative feature that can improve classification performance.

Formally, to incorporate the visual concepts, which are text string, into the image learning process, all the $N$ visual concepts in $P_{vc}$ are processed by the text encoder $g(\cdot)$ to obtain the feature matrix of concepts $\mathbf{H} \in \mathbb{R}^{N \times D}$. Then the image feature $\mathbf{z} \in \mathbb{R}^{1 \times D}$ acts as the query to perform cross-attention with textual features:
\begin{align}
\mathbf{a} & = Softmax(\frac{\mathbf{q} \mathbf{K}^T}{\sqrt{D}})\,, \label{eq:attention}\\
\mathbf{z}_{aux} & = \mathbf{a} \mathbf{V} \,,
\end{align}
where $\mathbf{q}=\mathbf{z}' \mathbf{W}_{q}, \mathbf{K}=\mathbf{H}' \mathbf{W}_k, \mathbf{V}=\mathbf{H}' \mathbf{W}_v$, and $\mathbf{z}'$ and $\mathbf{H}'$ are the image feature and concept features processed by LayerNorm, respectively. $\mathbf{W}_{q}$, $\mathbf{W}_{k}$, and $\mathbf{W}_{q}$ are the transformation matrices for query, key, and value , respectively. 
$\mathbf{a} =[a_1, a_2, ..., a_N]$ represents the attention score across $N$ concepts. Within this framework, the resulting fused feature $\mathbf{z}_{aux}$ possesses the semantic information of the query's corresponding class. Then $\mathbf{z}'_{aux}$, which is processed by the projection layer $\mathbf{W}_o$, passes through an auxiliary classifier head $\mathbf{h}_{aux}$, and the output is linearly fused with that of the solely image feature-based classifier $\mathbf{h}$ 
to obtain the final classification output.

This module quantifies the attention weights assigned by the image feature to each concept through cross-attention. Assigning more attention to image-relevant concepts is desirable. Therefore, to further encourage the module to focus on visual concepts corresponding to class of the image and to reduce interference from other irrelevant concepts, consequently enhancing the discriminability of the fused features, we design a loss function $\mathcal{L}_{attn}$ to constrain the attention scores, i.e.,
\begin{align}
    \mathcal{L}_{attn}=-\sum_{i \in S(c)} log(a_i) \,,
\end{align}
where $S(c)$ denotes the set of indices of visual concepts corresponding to class $c$, and $a_i$ is the $i$-th element of the attention vector (Equation~\ref{eq:attention}). Minimizing this negative log loss help increase attention scores for concepts belonging to the class of the input image and decrease all the other concepts irrelevant to the image class.

In summary, during training for each task, the overall loss is
\begin{align}
    \mathcal{L}=\alpha \mathcal{L}_{ce}+(1-\alpha) \mathcal{L}_{aux}+\lambda \mathcal{L}_{attn} ~.
\end{align}
The $\mathcal{L}_{ce}$ is the cross-entropy loss for the image classifier, while $\mathcal{L}_{aux} $ is for the auxiliary classifier. In addition, in order to mitigate forgetting during training, we construct pseudo-feature replay to simulate old class features and use them for training along with new data. Specifically, at the end of each learning task, a multivariate Gaussian distribution $\mathcal{N}(\mu, \Sigma)$  is constructed and saved for each class of this task as the pseudo-feature distribution, and the pseudo-features are sampled from these distributions while training in later tasks. During inference, the final prediction results is obtained by weighting the sum of the outputs of two classifier heads, as shown in \Cref{fig:method}.

\section{Experiments}


\subsection{Experimental Setup}

\noindent \textbf{Datasets:}
The proposed method was evaluated on multiple datasets across diverse domains, specifically in the medical and natural image domains. The medical datasets include Skin40~\cite{Skin40}, Skin8~\cite{Skin8}, and MedMNIST18~\cite{medmnist}. Skin40 is a subset of the SD-198 skin disease dataset, consisting of 40 classes. Skin8 is a balanced subset sampled from the 2019 ISIC Skin Challenge dataset, which has 8 classes in total. MedMNIST18 comprises two subdatasets, DermaMNIST and OrganCMNIST, which collectively encompass a total of 18 classes focused on dermatological and organ-related CT images. In addition, we used CUB200~\cite{CUB200}, a fine-grained bird dataset featuring 200 distinct classes, to explore the performance of our method in the natural image domain. The details of each dataset are summarized in \Cref{tab:dataset}. In the experiments, various CIL settings were adopted, i.e., 5-task and 10-task settings for Skin40, 4-task setting for Skin8,  6-task and 9-task settings for MedMNIST18,  and 5-task, 10-task and 20-task settings for Cub200. 

\begin{table}[!h]
    \centering
    \caption{Stats of each dataset.}
    \label{tab:dataset}
    \setlength{\tabcolsep}{5.5pt}
    \begin{tabular}{ccccc}
        \toprule
         Dataset & Modality & \#Classes & \#Train Set & \#Test Set \\
         \midrule
         Skin40~\cite{Skin40} & Dermoscopy & 40 & 2000 & 400  \\
         Skin8~\cite{Skin8} & Dermoscopy & 8 & 3555 & 705  \\
         MedMNIST18~\cite{medmnist} & Hybrid & 18 & 23377 & 10221  \\
         CUB200~\cite{CUB200} & Natural & 200 & 5994 & 5794 \\
         \bottomrule
    \end{tabular}

\end{table}

\noindent \textbf{Implementation Settings:} 
The pretrained CLIP ViT-B/16 \cite{CLIP} model is used as the default backbone. The foundation LLM for concept generation is Lingshu-7B~\cite{lingshu}, a multimodal language model specifically designed for the medical field. During training, AdamW optimizer is used for model training. The learning rate is set to 0.005 with cosine annealing learning rate schedule, and weight decay is 0.0001. The batch size is fixed to 32, and the number of training epochs is set to 20. In our method, the default number of concepts per class is set to 3, with a threshold value $\tau$ of 0.5 in the concept filtering mechanism. The coefficient for the loss function, $\alpha$, is set to 0.8, $\lambda$ is set to 0.6. Each experiment is run three times using different random seeds to ensure the robustness of the results.


\noindent \textbf{Metrics:} 
In this study, for the class-balanced dataset (Skin40), Last Accuracy ($\mathcal{A}_{Last}$) and Average Accuracy ($\mathcal{A}_{Avg}$) are employed as evaluation metrics. Last Accuracy denotes the model's accuracy across all learned classes after finishing all learning tasks, whereas Average Accuracy represents the mean of test accuracies observed at each learning stage. For the class imbalanced datasets (Skin8, MedMNIST18, CUB200), we report Last MCR ($\mathcal{M}_{Last}$) and Average MCR ($\mathcal{M}_{Avg}$), where MCR is mean class recall that is calculated by averaging the recall rate of each class.

\noindent \textbf{Baselines:}
Our method was compared with several CIL methods, including single-modal methods L2P\~cite{L2P}, DualPrompt~\cite{DualPrompt}, CODA-Prompt~\cite{CODA-Prompt}, and SLCA~\cite{CODA-Prompt}, and multi-modal methods PROOF~\cite{PROOF}, MoE4CIL~\cite{MoE-adapters}, Continual-CLIP~\cite{Continual-CLIP}, and CBM~\cite{CBM}. Among these, PROOF requires rehearsal of old samples, where we set the memory buffer size to 80 for the Skin40 and 5 per class for the other datasets. Furthermore, on medical image datasets, we also performed experiments involving unfreezing the last two layers of the image encoder during the first task training, with results designated as ``Ours*''. This modification was implemented to help domain adaptation to medical imaging, as the deeper layers of the image encoder capture domain-specific semantic features.

\begin{table*}[!ht]
    \centering
    \caption{Continual learning performance comparison on Skin40, Skin8 and MedMNIST18. ``Ours*'' represents the results of our method with the last two layers' tuning of the image encoder during the first task training. $^{\dagger}$ denotes the method has no standard deviation.}
    \label{tab:main_results}
    \setlength{\tabcolsep}{5pt}
    \begin{tabular}{cccccccccccc}
    \toprule
    \multirow{3}{*}{Method} &
      \multirow{3}{*}{} &
      \multicolumn{4}{c}{Skin40} &
      \multicolumn{2}{c}{Skin8} &
      \multicolumn{4}{c}{MedMNIST18} \\ \cmidrule{3-12} 
     &
       &
      \multicolumn{2}{c}{5-task} &
      \multicolumn{2}{c}{10-task} &
      \multicolumn{2}{c}{4-task} &
      \multicolumn{2}{c}{6-task} &
      \multicolumn{2}{c}{9-task} \\ \cmidrule(r){3-4} \cmidrule(r){5-6} \cmidrule(r){7-8} \cmidrule(r){9-10} \cmidrule(r){11-12} 
     &
       &
      $\mathcal{A}_{Avg}$ &
      $\mathcal{A}_{Last}$ &
      $\mathcal{A}_{Avg}$ &
      $\mathcal{A}_{Last}$ &
      $\mathcal{M}_{Avg}$ &
      $\mathcal{M}_{Last}$ &
      $\mathcal{M}_{Avg}$ &
      $\mathcal{M}_{Last}$ &
      $\mathcal{M}_{Avg}$ &
      $\mathcal{M}_{Last}$ \\ \midrule
    L2P~\cite{L2P}            &  & 66.57\std{0.97} & 43.92\std{0.63} & 58.69\std{0.78} & 37.25\std{1.25} & 49.75\std{1.02} & 27.73\std{0.92} & 54.89\std{1.55}       & 28.49\std{1.50}       & 38.22\std{1.34} & 20.14\std{1.54} \\
    DualPrompt~\cite{DualPrompt}     &  & 66.97\std{0.75} & 42.17\std{1.28} & 61.26\std{0.88} & 37.50\std{1.32} & 50.75\std{1.17} & 23.98\std{2.97} & 54.72\std{1.08}       & 27.50\std{1.86}       & 38.92\std{0.89} & 19.41\std{0.98} \\
    CODA-Prompt~\cite{CODA-Prompt}    &  & 62.37\std{0.34} & 40.17\std{0.88} & 54.84\std{1.06} & 30.25\std{2.38} & 48.32\std{0.47} & 19.37\std{2.06} & 55.97\std{2.19}       & 29.59\std{2.92}       & 38.17\std{1.81} & 18.01\std{0.69} \\
    Continual-CLIP$^{\dagger}$~\cite{Continual-CLIP} &  & 28.85 & 14.75 & 31.93 & 14.75 & 25.72 & 15.00 & 26.12 & 15.81 & 31.20 & 15.81 \\
    MoE4CIL~\cite{MoE-adapters}        &  & 66.09\std{1.55} & 41.33\std{1.38} & 52.44\std{0.78} & 23.75\std{1.52} & 47.65\std{0.35} & 22.72\std{0.92} &      38.83\std{1.50}       &      20.52\std{1.36}       &      43.78\std{0.63} &      21.64\std{0.81} \\
    PROOF~\cite{PROOF}          &  & 73.00\std{0.62} & 53.00\std{2.14} & 71.07\std{0.18} & 50.08\std{1.13} & 55.18\std{1.32} & 28.89\std{2.91} & 61.72\std{0.24}       & 51.93\std{1.58}       & 66.48\std{1.33} & 50.53\std{1.02} \\
    SLCA~\cite{SLCA}           &  & 73.73\std{0.54} & 55.92\std{1.01} & 66.51\std{} & 46.75\std{} & 59.17\std{0.96} & 31.87\std{1.52} & 60.58\std{1.50}       & 31.52\std{0.54}       & 48.38\std{0.76} & 29.75\std{0.49} \\ 
    \midrule
    \rowcolor{gray!20}
    Ours           &  &  73.80\std{1.60} & 63.67\std{1.53}  &  74.07\std{1.53} & 63.92\std{0.54} & 62.44\std{0.74} & 49.76\std{0.14} &  74.24\std{1.42}  &  63.29\std{1.78}  &  71.24\std{1.77}  &  62.48\std{0.61}  \\ 
    \rowcolor{gray!20}
    Ours*          &  &  74.51\std{1.50} & 64.33\std{1.01} & 74.95\std{0.88} & 66.17\std{0.84} & 65.79\std{1.69} & 51.55\std{1.55} &  77.67\std{1.83} & 67.45\std{0.45} & 76.16\std{1.47} & 66.47\std{1.41}\\
    \bottomrule
    \end{tabular}
\end{table*}

\begin{table*}[!ht]
    \centering
    \caption{Continual learning performance comparison on CUB200 in 5-, 10- and 20-task settings.}
    \label{tab:cub}
    \begin{tabular}{ccccccc}
    \toprule
    \multirow{3}{*}{Method} & \multicolumn{6}{c}{CUB200} \\ \cmidrule{2-7} 
    &   \multicolumn{2}{c}{5-task} & \multicolumn{2}{c}{10-task} & \multicolumn{2}{c}{20-task} \\ \cmidrule(r){2-3} \cmidrule(r){4-5} \cmidrule(r){6-7}  
    &   $\mathcal{M}_{Avg}$ & $\mathcal{M}_{Last}$ & $\mathcal{M}_{Avg}$ & $\mathcal{M}_{Last}$ & $\mathcal{M}_{Avg}$ & $\mathcal{M}_{Last}$ \\ \midrule
    L2P~\cite{L2P}              & 78.57\std{0.27} & 68.59\std{0.31} & 74.89\std{0.11} & 63.05\std{0.41} & 67.90\std{0.58} & 53.71\std{0.59} \\
    DualPrompt~\cite{DualPrompt}       & 78.53\std{0.17} & 68.44\std{0.17} & 75.20\std{0.37} & 63.09\std{0.42} & 69.31\std{0.42} & 55.48\std{0.84} \\
    CODA-Prompt~\cite{CODA-Prompt}      & 76.92\std{0.13} & 66.27\std{0.19} & 72.66\std{0.80} & 59.76\std{0.27} & 68.96\std{0.77} & 53.62\std{1.57} \\
    Continual-CLIP~\cite{Continual-CLIP}   & 59.89 & 48.53 & 61.86 & 48.53 & 63.04 & 48.53 \\
    MoE4CIL~\cite{MoE-adapters}          & 72.05\std{0.40} & 59.08\std{0.57} & 70.48\std{1.05} & 55.17\std{1.57} & 66.26\std{0.92} & 50.77\std{0.02} \\
    PROOF~\cite{PROOF}            & 81.78\std{0.10} & 72.07\std{0.08} & 82.43\std{0.11} & 72.45\std{0.38} & 82.64\std{0.23} & 71.70\std{0.38} \\
    CBM~\cite{CBM} & 82.29\std{0.65} & 77.06\std{0.89} & 83.82\std{0.21} & 76.65\std{0.58} & 82.37\std{0.18} & 76.86\std{0.49} \\
    SLCA~\cite{SLCA}             & 81.90\std{0.09} & 73.21\std{0.15} & 81.02\std{0.09} & 70.27\std{0.30} & 81.02\std{0.15} & 68.35\std{0.49} \\ 
    \midrule
    \rowcolor{gray!20}
    Ours        & 84.28\std{0.35} & 78.35\std{0.31} & 85.00\std{1.05} & 78.24\std{0.26}  & 84.57\std{0.54} &  77.73\std{0.62}      \\ 
    \bottomrule
    \end{tabular}
\end{table*}

\begin{figure*}[!t]
    \centering
    \begin{subfigure}[t]{0.3\linewidth}
        \centering
        \includegraphics[width=\linewidth]{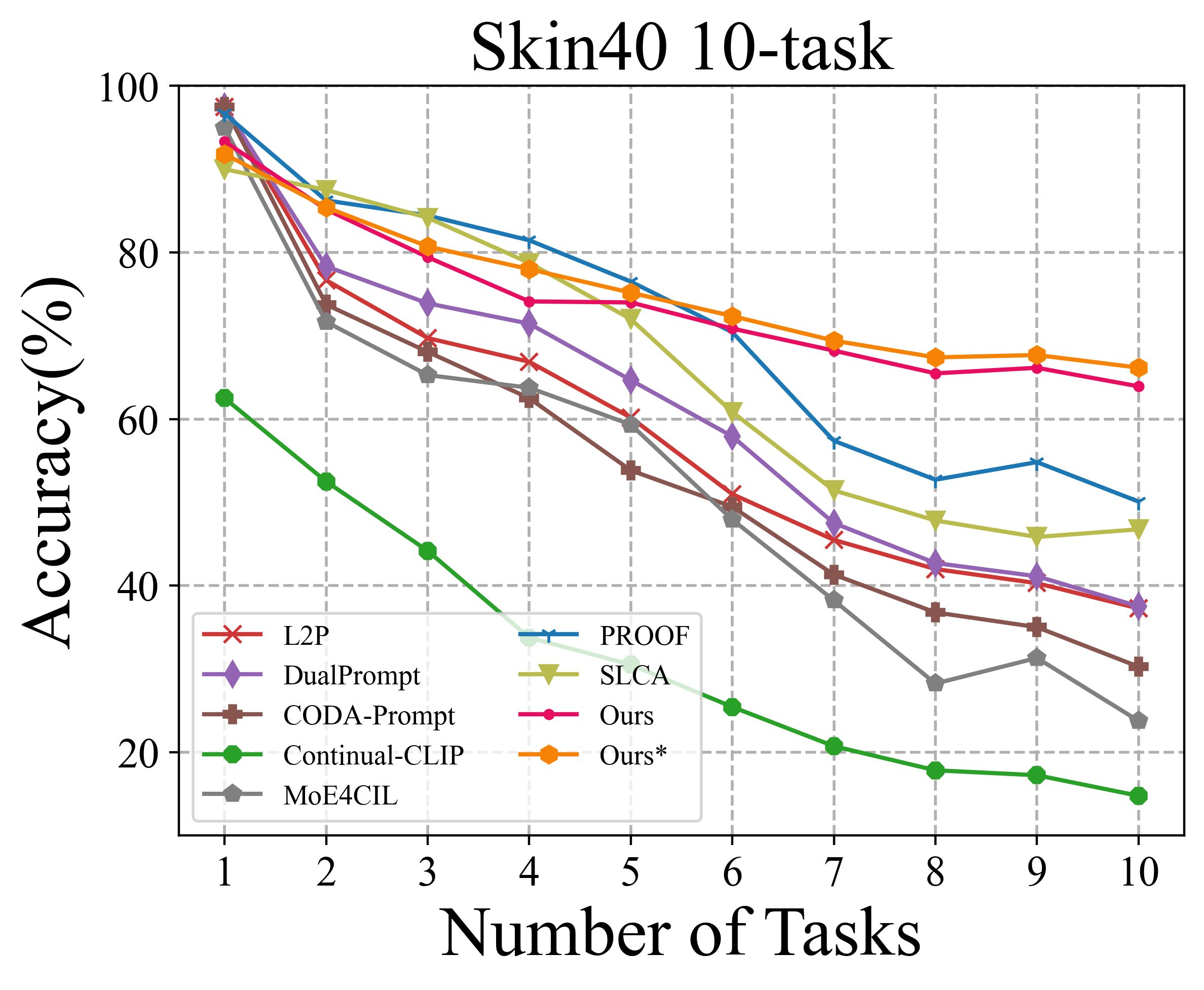}
    \end{subfigure}
    \begin{subfigure}[t]{0.3\linewidth}
        \centering
        \includegraphics[width=\linewidth]{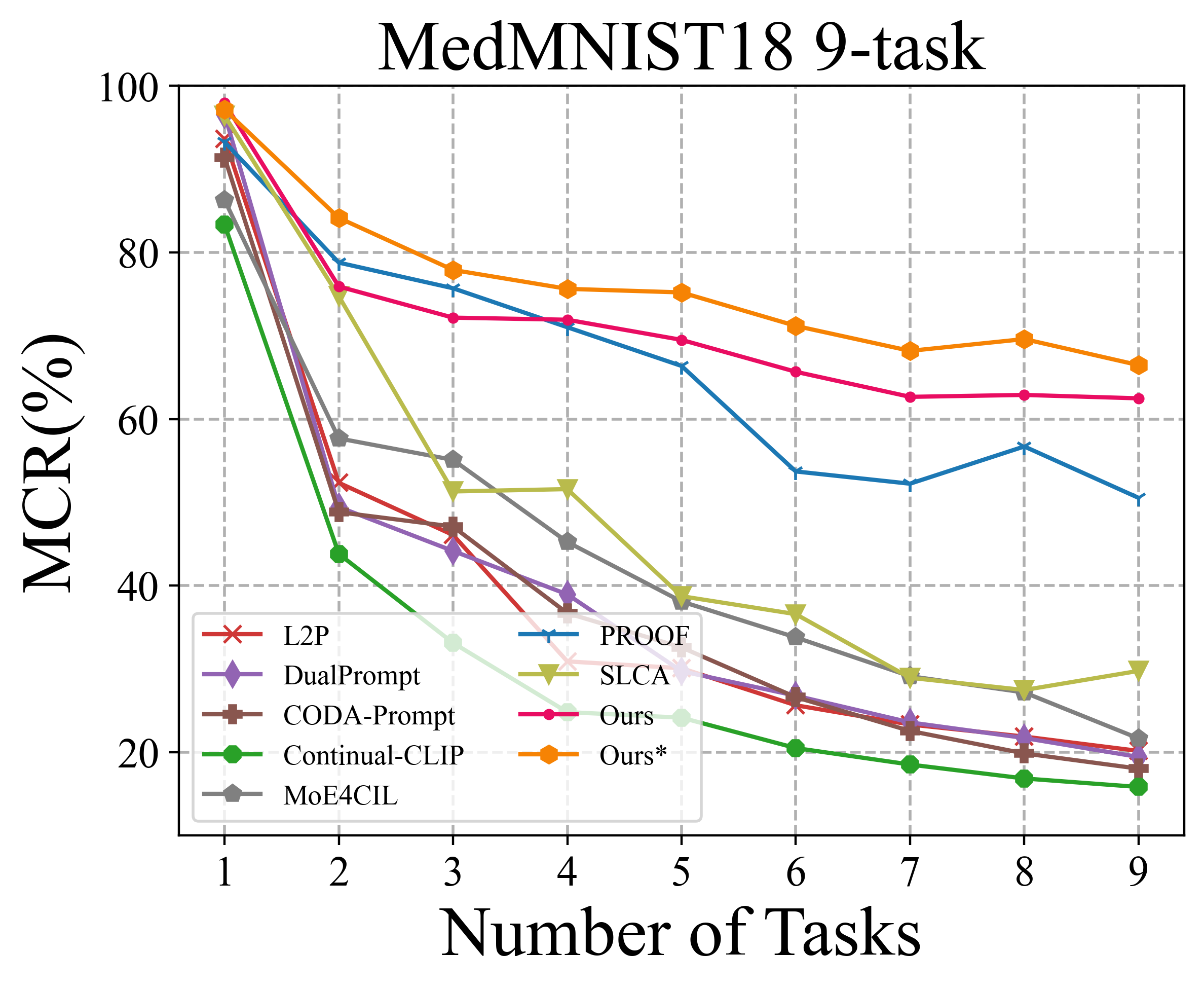}
    \end{subfigure}
    \begin{subfigure}[t]{0.3\linewidth}
        \centering
        \includegraphics[width=\linewidth]{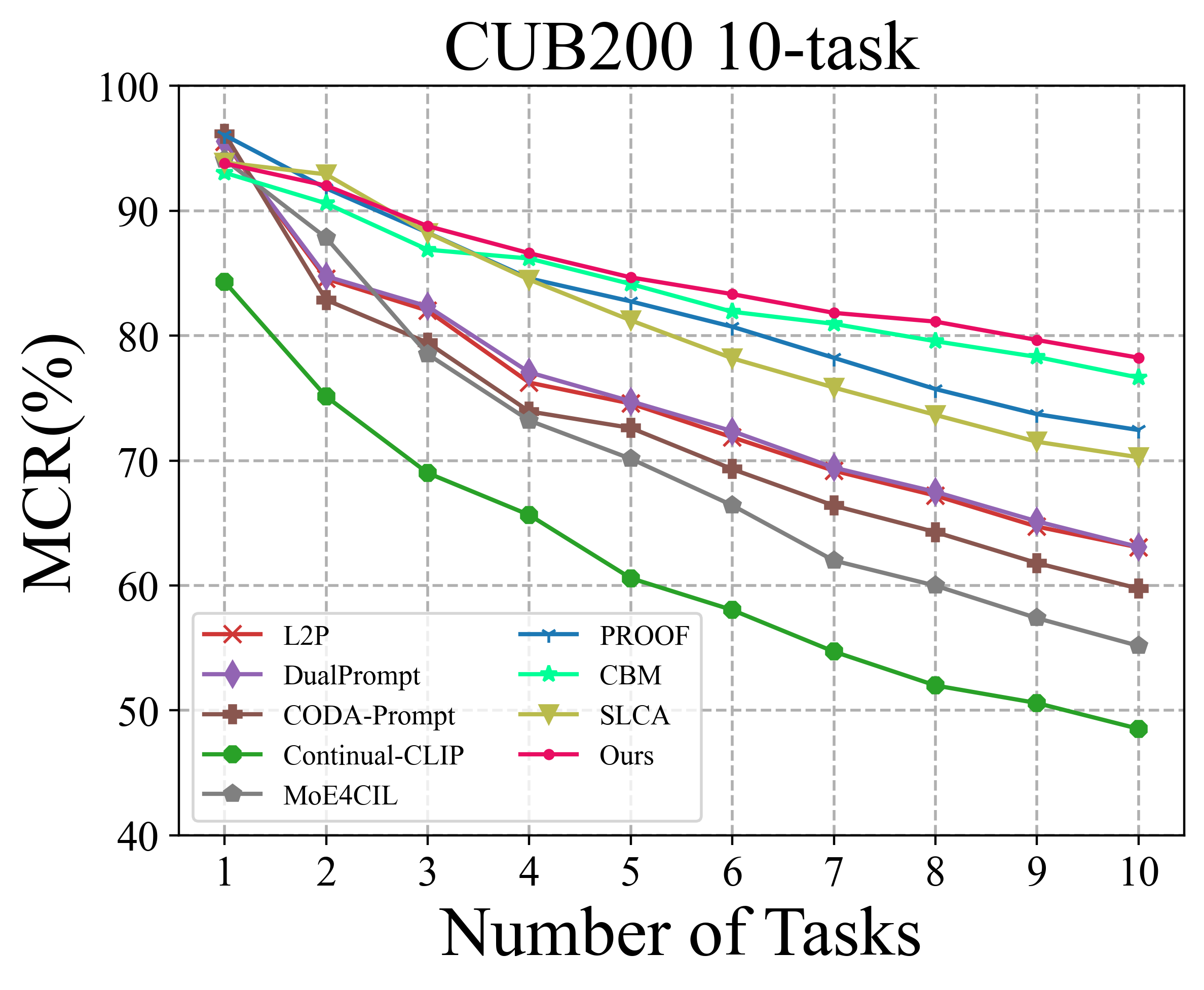}
    \end{subfigure}
    \caption{The performance curves of different methods over the whole CIL process on Skin40, MedMNIST18 and CUB200.}
    \label{fig:curves}

\end{figure*}

\subsection{Effectiveness Evaluation}
The effectiveness of our method was evaluated by comparing with strong baselines on both medical and natura images datasets. 
As shown in \Cref{tab:main_results}, our method consistently achieves superior performance on a range of benchmarks. For example, on Skin40, our method recorded last accuracy of 64.33\% and 66.17\% for the 5-task and 10-task settings, respectively, significantly outperforming competing methods by at least 7\%. Besides, on MedMNIST18 under the 9-task setting, our method surpassed all prompt-based counterparts, such as L2P, DualPrompt, and CODA-Prompt, by a margin of at least 30\% in last MCR. Furthermore, it is clear that our method outperformed PROOF, which also utilizes cross-modal attention fusion, by at least 10\% in last accuracy or last MCR across three datasets, demonstrating the superiority of our attention fusion mechanism. Furthermore, we verified that unfreezing the last two layers of the image encoder during the first task training facilitates its adaptation to the medical imaging domain, leading to a further performance gain as shown in the last row of \Cref{tab:main_results} (\textit{i.e.}, Ours*).


In addition to the superior performance on the medical image dataset, our method also achieves the best results on CUB200. Particularly, the runner-up method CBM also leverages LLMs to generate class descriptions to help mitigate forgetting. Our method outperforms CBM by 1.59\% in last MCR under the 10-task setting, further supporting the superiority of our visual concepts. In addition, as expected, over the whole incremental learning process,
our method significantly outperforms other baselines by a large margin (\Cref{fig:curves}), further confirming the efficacy and superiority of our method.

\subsection{Ablation Study}

\begin{table}[!t]
    \centering
    \caption{Ablation study of our method on Skin40 under 10-task setting. $\Delta_{Last}$ denotes the change in $\mathcal{A}_{Last}$ relative to the previous row. ``Concept Branch'' means the auxiliary branch of our method. ``First Task Adaptation'' means unfreezing the last two layers of the image encoder during first task training.}
    \label{tab:ablation}
    \setlength{\tabcolsep}{10pt}
    \begin{tabular}{lccc}
    \toprule
    & $\mathcal{A}_{Avg}$ & $\mathcal{A}_{Last}$ & $\Delta_{Last}$ \\
    \midrule
    Baseline: Feature Replay   & 71.00\std{0.51}      & 60.45\std{0.47}  & -     \\
    w/ Concept Branch                  & 73.01\std{0.26}      & 62.13\std{0.38}  & +1.68      \\
    w/ $\mathcal{L}_{attn}$              & 74.07\std{0.33}      & 63.92\std{0.54}  & +1.79      \\
    w/ First Task Adaptation             & 74.95\std{0.71}      & 66.17\std{0.84}  & +2.25      \\
    \bottomrule
    \end{tabular}
\end{table}

We conducted an ablation study to investigate the impact of each module in our proposed method, with results summarized in \Cref{tab:ablation}. The baseline, which relied solely on Gaussian distribution for pseudo-feature replay in continual learning, achieved a last accuracy of 60.45\% on the Skin40 in 10-task setting. The second row shows that introducing our proposed visual concept pool, coupled with the integration of cross-attention to obtain fused features for the auxiliary classification head, yielded an improvement of 1.68\% in last accuracy, which proves the effectiveness of the visual concept branch. Furthermore, by incorporating the attention loss $\mathcal{L}_{attn}$ based on row 2, the model is constrained to focus more on the features of image-relevant concepts, resulting in a further enhancement of 1.79\% in last accuracy. Finally, unfreezing the last two layers of the image encoder in the first task training phase, aimed at adapting the model to the downstream medical domain, achieves a last accuracy that is 2.25\% higher than the result without adaptation. Collectively, these results demonstrate the effectiveness and significance of each module of our method in augmenting model performance.

\subsection{Sensitivity Analysis}

\begin{figure}[!t]
    \centering
    \begin{subfigure}[t]{0.49\linewidth}
        \centering
        \includegraphics[width=\linewidth]{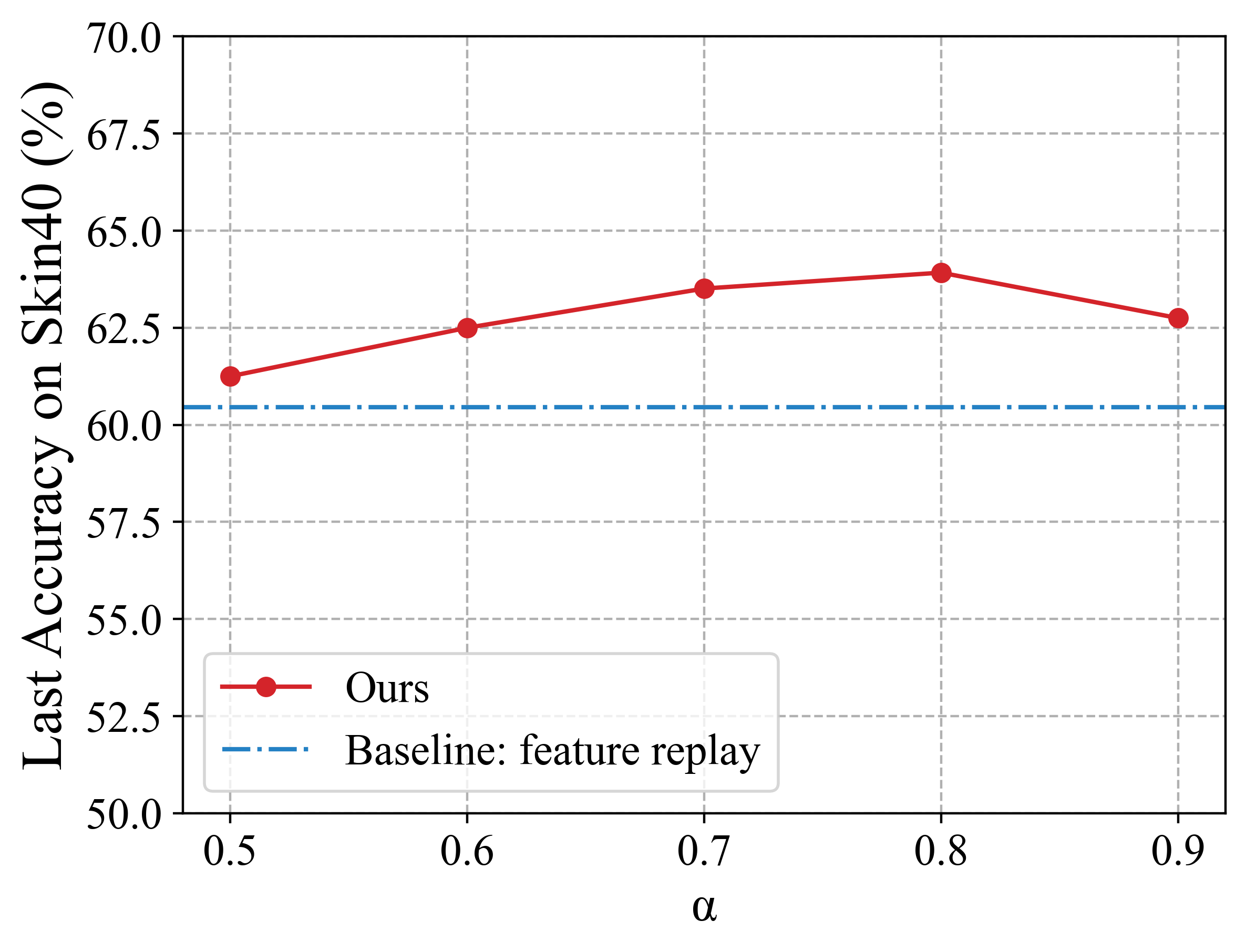}
    \end{subfigure}
    \begin{subfigure}[t]{0.49\linewidth}
        \centering
        \includegraphics[width=\linewidth]{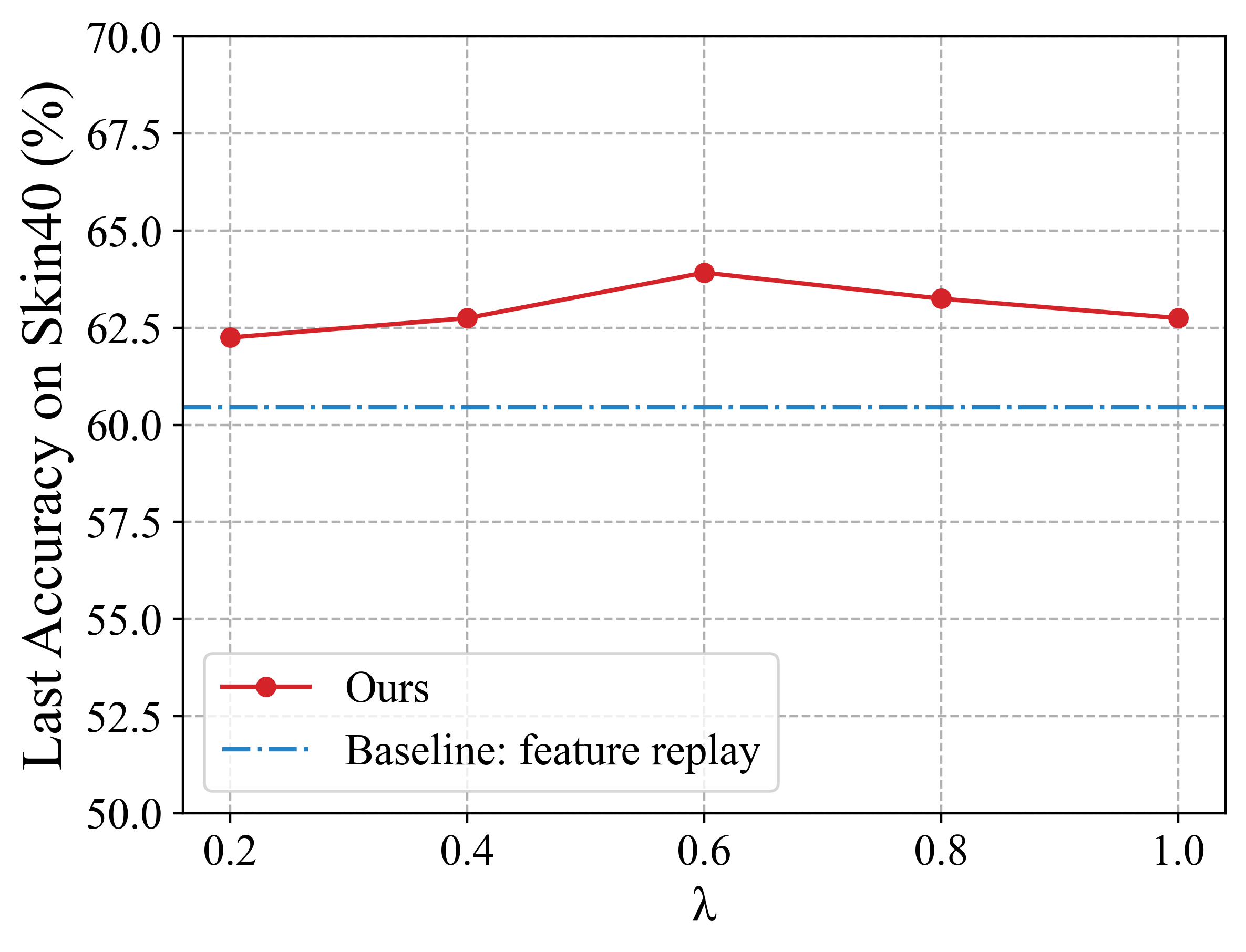}
    \end{subfigure}
    \caption{Sensitivity analysis of the loss coefficients $\alpha$ and $\lambda$ on Skin40 under the 10-task setting.}
    \label{fig:alpha-lambda}
\end{figure}

\noindent \textbf{Analysis of $\alpha$:} 
Sensitivity analysis was performed on the hyperparameter $\alpha$, as shown on the left of \Cref{fig:alpha-lambda}. Within the range of 0.5 to 0.9, our method is largely insensitive to $\alpha$, achieving a better last accuracy on Skin40 under the 10-task setting than only using feature replay. In particular, at $\alpha=0.8$, our method achieves a peak performance of 63.92\%. This shows that the auxiliary branch in our method can consistently enhance the classification performance of the image branch through a weighted combination.

\begin{figure*}[!t]
    \centering
    \includegraphics[width=\textwidth]{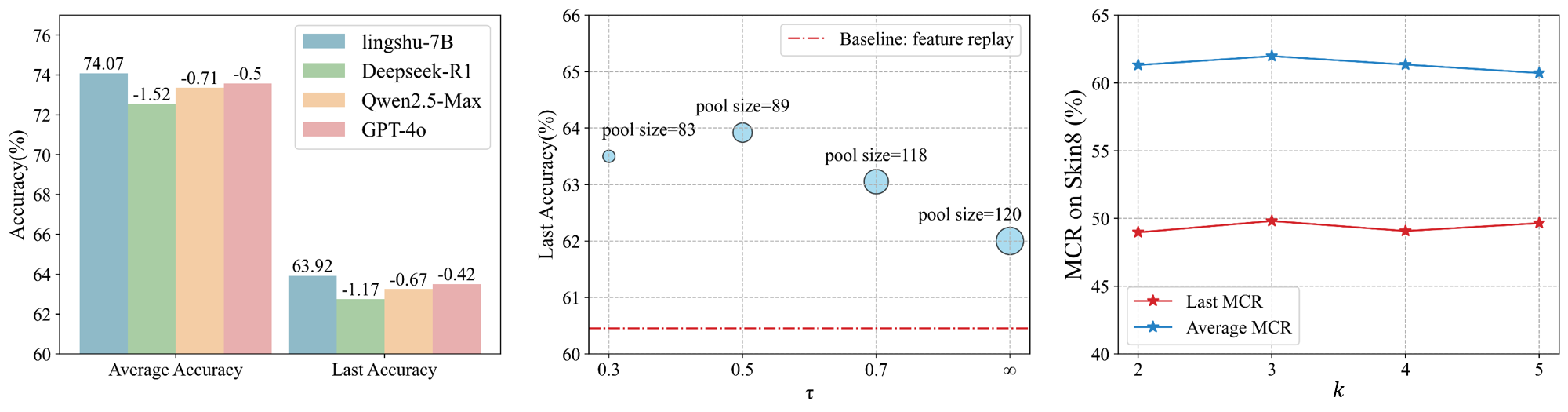}
    \caption{Analysis of visual concepts generation. Left: performance of our method with different LLMs on Skin40 under the 10-task setting. Middle: analysis of the filtering mechanism threshold $\tau$ on Skin40 under the 10-task setting. The size of the circle represents the size of the concept pool. Right: analysis of the number of concepts per class on Skin8.}
    \label{fig:llm-tau-k}
\end{figure*}

\begin{figure}[!t]
    \centering
    \includegraphics[width=\linewidth]{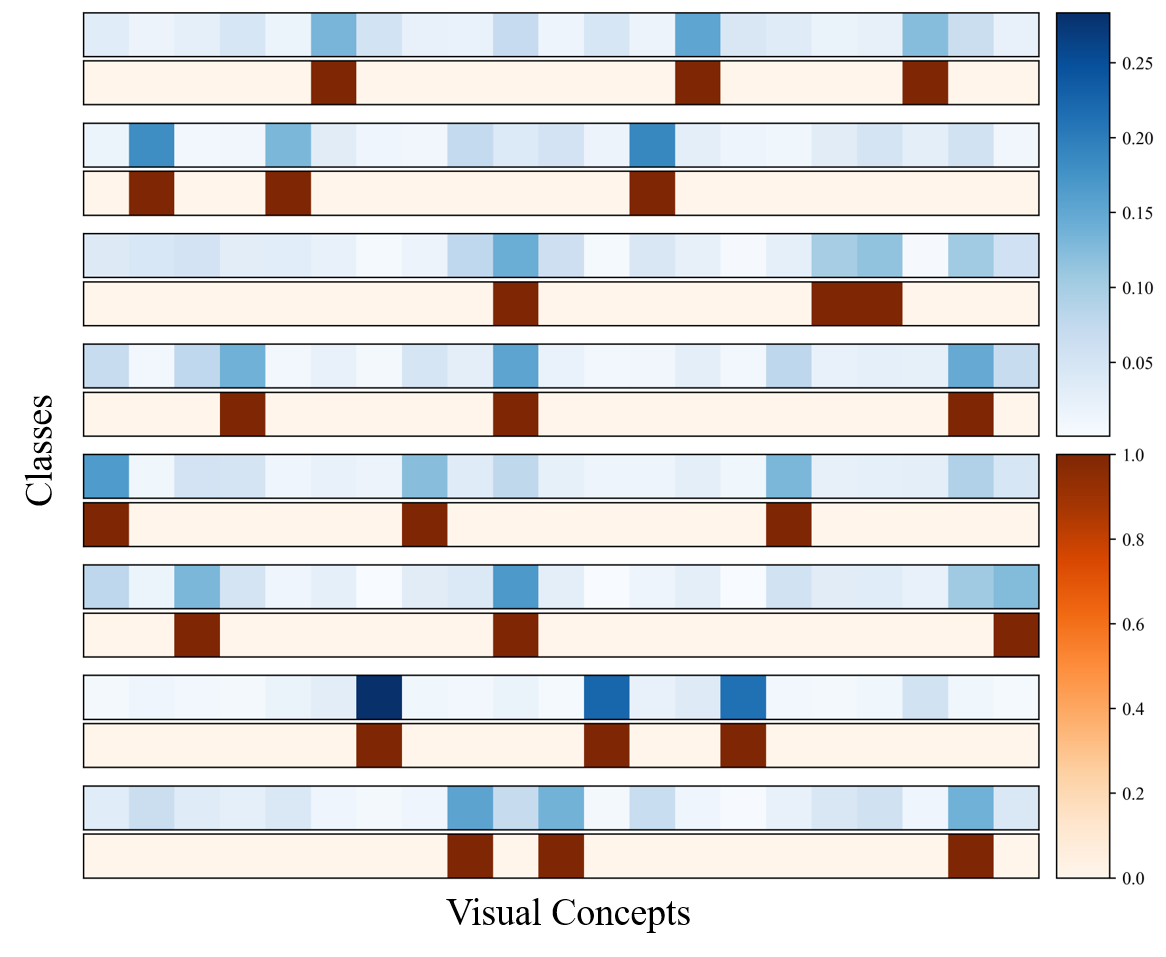}
    \caption{Visualization of our cross-attention scores on Skin8. Rows represent distinct classes, while columns denote different visual concepts. Each row comprises two groups: blue one indicates the mean attention scores, and the orange one corresponds to the ground-truth concepts.}
    \label{fig:attn}
\end{figure}

\noindent \textbf{Analysis of $\lambda$:}
We performed a sensitivity analysis on $\lambda$ to assess its impact on the performance of our method. The findings reveal that our method demonstrates stable performance across a wide range of $\lambda$ values (\Cref{fig:alpha-lambda}, right). The optimal value of $\lambda$ is identified at 0.6, where the model achieves the highest last accuracy. Notably, when $\lambda > 0.6$, the performance experiences a slight degradation. This occurs because an excessively strong attention loss constraint dominates the training process, weakening the impact of the classification loss. Overall, our method is largely insensitive to the value choice of  $\lambda$. 

\subsection{Analysis of Visual Concepts Generation}

\noindent \textbf{Different LLMs:}
To verify the impact of different concept sources in our method, we performed experiments on diverse LLMs, with results illustrated in  \Cref{fig:llm-tau-k} (left). It shows that our method can also achieve impressive results on DeepSeek-R1~\cite{deepseek-r1}, Qwen2.5-Max~\cite{qwen2.5}, and GPT-4o~\cite{gpt4}, which validates our method's cross-LLM robustness. Notably, employing the medical LLM lingshu-7B can achieve slightly better results, as it yields more specialized concepts due to its expert-level medical knowledge.

\noindent \textbf{Different threshold $\tau$:}
We conducted an analysis on the threshold $\tau$ of the similarity-based concept filtering mechanism in our method. The size of the resulting visual concept pool and the corresponding model performance under different values of $\tau$ are illustrated in \Cref{fig:llm-tau-k} (middle). Our method achieves the highest last accuracy of 63.92 on Skin40 when $\tau=0.5$, with the concept pool containing 89 distinct concepts. When $\tau$ is larger than or smaller than 0.5, the performance slightly degrades. In the case where $\tau=\infty$, which means that no filtering mechanism is applied, the performance drops by 1.92\% compared to that of $\tau = 0.5$. This is because of the presence of highly similar concepts in the concept pool, which introduces conflicting supervision signals during attention loss optimization. In general, the results demonstrate the effectiveness of the filtering mechanism in eliminating redundancy and improving learning consistency.

\noindent \textbf{Different numbers of concepts per class $k$:}
To further demonstrate the robustness of the concept generation module, we conducted sensitivity analysis on the number $k$ of concepts per class, with results shown in \Cref{fig:llm-tau-k} (right). When $k$ varies from 2 to 5, the performance of our method remains stable, with accuracy fluctuations  within 1\%. 

\subsection{Visualization Analysis}
To validate the efficacy of our cross-modal attention mechanism, we visualize the attention scores, with results on the Skin8 test set shown in \Cref{fig:attn}. Rows correspond to classes, and columns represent different concepts. Each row contains two groups of values: the first group (blue) shows the average attention scores of all samples for the class, while the second group (orange) indicates the ground-truth concepts ($k=3$) associated with the class. As the figure shows, the attention score distribution over concepts for each class aligns well with the ground truth,  and no forgetting issue is observed. This qualitative visualization demonstrates that our attention loss function effectively enforces the model to focus on input-relevant concepts during CIL process, validating its capability in establishing precise cross-modal associations.

\section{Conclusion}
In this paper, we propose a visual concept-guided continual learning framework for disease classification. We generate visual concepts for novel classes leveraging LLMs. To mitigate concepts conflict, we maintain a concept pool and dynamically update it via a filtering mechanism. Besides, a cross-modal attention mechanism is introduced to enable image-text interaction, where attention scores and the proposed attention loss integrate the most relevant concept features into the output fused feature. The fused feature is processed through an auxiliary classifier, and the outputs can augment the final classification results via a weighted sum with the outputs of the image branch. Extensive experiments show that our approach achieves state-of-the-art performance across multiple medical and natural image datasets. Future work will focus on extracting local image features and aligning them with concept texts to further leverage the guiding role of concepts.

\small
\bibliographystyle{IEEEtran}
\bibliography{main}






\end{document}